\documentclass[letterpaper, 10 pt, conference]{ieeeconf}  

\IEEEoverridecommandlockouts                              

\usepackage{graphics} 
\usepackage{epsfig} 
\usepackage{mathptmx} 
\usepackage{times} 
\usepackage{amsmath} 
\usepackage{bm}
\usepackage{amssymb}  
\usepackage{color}
\usepackage{subfig}
\newtheorem{theorem}{Theorem}
\newtheorem{lemma}{Lemma}

\usepackage[ruled,lined,boxed,commentsnumbered]{algorithm2e}

\title{\LARGE \bf Analysis of Obstacle based Probabilistic RoadMap Method using Geometric Probability}

\author{Titas Bera$^{1}$, M. Seetharama Bhat$^{2}$, and Debasish Ghose$^{3}$
\thanks{$^{1}$Titas Bera is a research Scholar at Indian Institute of Science, 
   Bangalore, India 
        {\tt\small titasbera@aero.iisc.ernet.in}}%
\thanks{$^{3}$M. Seetharama Bhat is a Professor in the Aerospace Engineering Department, India
        {\tt\small msbdcl@aero.iisc.ernet.in}}%
\thanks{$^{3}$Debasish Ghose is a Professor in the Aerospace Engineering Department, India
        {\tt\small dghose@aero.iisc.ernet.in}}%
}

\begin{document}

\maketitle
\thispagestyle{empty}
\pagestyle{empty}

\begin{abstract}                
Sampling based planners have been successful in robot motion planning, with many degrees of freedom, but still remain ineffective in the presence of narrow passages within the configuration space. There exist several heuristics, which generate samples in the critical regions and improve the efficiency of probabilistic roadmap planners. In this paper, we present an evaluation of success probability of one such heuristic method, called obstacle based probabilistic roadmap planners or OBPRM, using geometric probability theory. The result indicates that the probability of success of generating free sample points around the surface of the $n$ dimensional configuration space obstacle is directly proportional to the surface area of the obstacles.
\end{abstract}


\section{Introduction}
An indispensable concept of modern robot motion planning theory is
the concept of configuration space. In these cases, the problem is to
find a path from an initial point to a goal
point in the configuration space. However, the explicit mapping
from workspace obstacles to configuration space is, in general
difficult. Early studies show that the basic version of motion planning
problem is PSPACE-complete and the best exact deterministic
algorithm known is exponential with the dimension of the
configuration space \cite{perez}, \cite{reif}. 

Since the mid-nineties, in order to break the curse of dimensionality, random sampling based approaches are introduced. Two such popular sampling based approaches are Probabilistic Roadmap Method or PRM, introduced by Kavraki et al. \cite{kavraki}, and Rapidly Exploring Random Tree or RRT, introduced by Lavalle \cite{lavalle}. These sampling based algorithms work in two phases. In the processing phase, a graph/tree structure is created, where nodes represent configurations and any edge between two nodes indicates that there exists a local planner which can connect the two configurations. In the subsequent query phase, a graph search algorithm tries to find a path
between any two given configurations. Because of various types of sampling scheme and local planners, there exist many variants of sampling based planners. The implementation of these algorithms is in general quite simple. For details, see \cite{choset} and for an excellent survey see \cite{carpin}. The implementation of these algorithms is usually quite simple. The price to pay is algorithmic completeness. Algorithms based on randomly generated samples aims for probabilistic completeness. Despite the success of sampling based path planners, motion planning in high dimensional configuration space is still difficult. For PRM like planners, one such difficulty arises when
configuration space possesses narrow passages. Several heuristic sampling strategies can remove this difficulty to a large extent, but a satisfactory answer remains elusive.

In this paper, we investigate one such heuristic, originally developed by Amato et al. \cite{amato}, called obstacle based roadmap method or OBPRM. The method relies on uniformly generating many sample points on the obstacle surface to build a graph structure in a probable narrow passage region within the configuration space. The method selects a point within the obstacle and shoots line segment in random directions. Then using a binary search method, it tries to find sample points within the free configuration space within a finite number of iterations, . We calculate the success probability of the OBPRM algorithm using the theory of geometric probability. These applications of various theoretical results of geometric probability are new in the literature.

This paper is organized as follows. In Section \ref{Problem formulation} we formulate the motion planning problem. Section \ref{prm_narrow} describes the narrow passage problem and various heuristics to solve the problem. In the following Section \ref{Cspace Obstacles}, we show how to model configuration space obstacles and its relationship with corresponding workspace obstacles in terms of feature complexity. In Section \ref{obprm_algo} we briefly describe OBPRM algorithm. Section \ref{def}, contains various definitions and theorems of geometric probability theory which are required to do probabilistic analysis of OBPRM. In Section \ref{analysis} we calculate the success probability of OBPRM algorithm. Secion \ref{Results} contains the results that supports the theoretical claim and finally in Section \ref{Conclusion} we conclude.

\section{Motion Planning Problem Formulation}\label{Problem formulation}
The configuration of a robot with $n$ degrees of freedom can be represented as a point in an $n$-dimensional space, called the configuration space $\mathbb{C}$, which is locally like the $n-$dimensional euclidian space $\mathbb{R}^n$. A configuration $q$ is free if the robot placed at $q$ does not collide with the obstacles or with itself. We define the free space $\mathbb{C}_{free}$ to be the set of all free configurations in $\mathbb{C}$. Assume that there are a finite number of obstacles in the workspace or physical space, which are closed bounded sets $\mathbb{O}_i$, $i=1,2,\ldots,m$ and we can fairly assume that they are pairwise disjoint. Let the starting configuration be $q_{start}\in\mathbb{C}_{free}$ and the final configuration be $q_{goal}\in\mathbb{C}_{free}$. For convergence issues, we define a rather general goal subset $C_{goal}$, than a specific point. Clearly $q_{free}\in{C_{goal}}$. The motion planning problem is to find a path connecting $q_{start}$ with $C_{goal}$, that is, a continuous function $f:[0,1]\rightarrow\mathbb{C}_{free}$ such that $f(0)=q_{start}$ and $f(1)\in{C_{goal}}$. We assume the existence of appropriate collision checker function or boolean predicate for every instance of the problem. In related literature, the configuration space $\mathbb{C}$ is frequently substituted with the state space $\mathcal{X}$, which includes both the degrees of freedom and their derivatives. The constraints can be non-holonomic and/or differential. 


\section{PRM and Narrow Passage Problem}\label{prm_narrow}
Probabilistic road-map methods (PRM) solve motion planning problems that do not involve dynamics of the robot or have
negligible dynamics. A classic multi-query PRM planner proceeds in two stages. In the first stage, it randomly chooses samples from $\mathbb{X}_{free}$, called milestones according to a sampling scheme. It then uses these milestones as nodes to construct a graph, called a road-map, by adding an edge between every $k$ pair of milestones that can be connected via a simple collision-free path, typically, a straight-line segment \cite{kavaraki_2}. After the road-map has been constructed, multiple queries can be answered quickly in the second stage. The planner first finds two milestones in the road-map, such that $x_{start}$ and $x_{goal}$ can be connected to these nodes and subsequently search for a path between $x_{start}$ and $x_{goal}$.

If the configuration space possesses a narrow passage then, to capture the connectivity of $\mathbb{X}_{free}$, it is essential to sample milestones in narrow passages. This, however, is difficult, because of small volumes of narrow passages. Uniform distribution may not work well when the dispersion of the samples is higher than the narrow passage volumes. Intuitively, one should sample more densely near obstacle boundaries because points in narrow passages lie close to obstacles. A method called Gaussian sampler \cite{boor} is a simple and efficient algorithm that uses
this idea. However, in some cases, many points near the obstacle boundaries lie far away from narrow passages and do not help in improving the connectivity of road-maps.

Another heuristic scheme for the problem of narrow passage is the bridge test \cite{Zheng}. Here when a sample lies
within an obstacle, one uses this information to build a bridge whose two end points lie in the obstacle while the mid point lies in $\mathbb{X}_{free}$. This method, although it requires a high computational time (because of more number of collision checking), can be very effective. Another approach called OBPRM \cite{amato}, is an obstacle based sampling method where random rays are cast from obstacles and using binary search one looks for collision free points near obstacle boundaries. Other geometric approaches also exist but those are expensive to implement in high-dimensional configuration spaces. Very recently another approach known as Toggle PRM, which builds a graph structure not only within $\mathbb{X}_{free}$, but also within configuration space obstacle, has gained popularity. It then uses both the graph information to generate samples within the narrow passage. For details see, \cite{choset}.

\section{Configuration Space Obstacles}\label{Cspace Obstacles}
In this section we will try to find face complexity of configuration space obstacles and also try to establish the relationship between workspace obstacle's surface area to the corresponding configuration space obstacle's surface area.
We define workspace as $W$ and either $W \in \mathbb{R}^2$ or $W \in \mathbb{R}^3$ for all practical robot motion planning problems. Obstacles can be considered closed sets in $W$. For simplicity we consider obstacles are rigid bodies. Each obstacle in the workspace can be represented as unions or intersections of a finite number of geometric primitives. In case of planar convex obstacle, boundary can be represented as a set of vertices and edges, and for a solid representation of obstacles, one can think of intersections of finite number of half planes. A convex polygonal obstacle therefore expressed as,
\begin{equation}
\mathbb{O} = H_1 \cap H_2 \cap \ldots \cap H_n 
\end{equation}
where, $H_i, 1 \leq i\leq n$, are halfplanes. For nonconvex polygons, it can be expressed as a finite union of convex obstacles. That is,
\begin{equation}
 \mathbb{O} = {\mathbb{O}}_1 \cup {\mathbb{O}_2} \cup \ldots \cup {\mathbb{O}_m}
\end{equation}
When $W \in \mathbb{R}^3$, arrangements of half space primitives can be used to define convex polyhedrons. The boundary representation consists of vertices, edges, and faces. Once again for non-convex polyhedron, it can be represented as a finite union of convex polyhedrons. We also assume that 
obstacles represented in $\mathbb{R}^2$ or $\mathbb{R}^3$ as semi-algebraic sets, can be well approximated using a finite number of half planes or half spaces. 


Now, suppose $W \in \mathbb{R}^2$ or $W \in \mathbb{R}^3$ contains an obstacle region, $\mathbb{O} \subset W$. Assume that a rigid robot is defined as $A$. Assume both the $A$ and $\mathbb{O}$ are expressed as semialgebraic models. Let $q \in \mathbb{C}$ denote a configuration of $A$ then the obstacle region $\mathbb{C}_{obs} \subset \mathbb{C}$ is defined as
\begin{equation}
 {\mathbb{C}}_{obs} = \{q\in \mathbb{C} | A(q) \cap \mathbb{O} \neq \emptyset\}
\end{equation}
which is the set of all configurations, $q$, at which $A(q)$, the transformed robot, intersects the obstacle region, $\mathbb{O}$. Since $\mathbb{O}$ and $A(q)$ are closed sets in $W$, the obstacle region is a closed set in $\mathbb{C}$.

The leftover configurations are called the free space, which is defined and denoted as $\mathbb{C}_{free} = \mathbb{C} \setminus \mathbb{C}_{obs}$. Since $\mathbb{C}$ is a topological space and $\mathbb{C}_{obs}$ is closed, $\mathbb{C}_{free}$ must be an open set. This implies that the robot can come arbitrarily close to the obstacles while remaining in free space. 
\subsection{Modelling of $\mathbb{C}_{obs}$: The Translational Case}
The simple case for characterizing $\mathbb{C}_{obs}$ is when $\mathbb{C} = \mathbb{R}^n$ for $n = 1,2 ,3$, and the robot is a rigid body that is restricted to translation only. Under these conditions, $\mathbb{C}_{obs}$ can be expressed as a type of convolution. For any two sets $X,Y \subset \mathbb{R}^n$, let their Minkowski sum be defined as 
\begin{equation}
X \oplus Y = \{x - y \in \mathbb{R}^n | x \in X ~ \text{and} ~ y \in Y\} 
\end{equation}
Let us assume that the configuration space has dimension $d$. This means we are assuming the robot has $d$ degrees of freedom. The set of all robot placements in which a specific robot feature, that is, vertex, edge, face is in contact with a specific obstacle feature is an $d-1$ dimensional subset, or hypersurfaces, in the $d$ dimensional configuration space. Each combination of robot feature and an obstacle feature defines such a hypersurface. If the total number of features of the robot and the obstacle are $m$ and $n$, respectively, then the total number of hypersurfaces equals $\mathbb{O}(mn)$. 
\subsection{Modeling of $\mathbb{C}_{obs}$: General Case}
Let $N$ be a set of hyperplanes in $\mathbb{R}^n$, where $n = 2,3$. The arrangement $G(N)$ formed by $N$ is defined to be the natural partition of $\mathbb{R}^n$ into convex regions or faces of varying dimensions along with the adjancecies among them. These convex regions are $n$-cells or $n$-faces of $G(N)$. Let us assume that this arrangement is simple or nondegenrate. This means the intersections of any $j > 0$ hyperplanes in $N$ is $(n-j)$ dimensional. The total number of faces of all dimensions is denoted as $|G(N)|$ which is called the size of the partition. Now we state the following lemma, that can be found in ch.6, \cite{mulmulle}.
\lemma{For a fixed $n$, $|G(N)| = O(l^n$), where $l$ is the size of $N$.}

Using this lemma, we can say that if the obstacle is represented as arrangement of hyperplanes then the face complexity of an obstacle in workspace is $O(l^n)$. Now let us assume that the configuration space $C$ has dimension $d$. The set of all robot placements in which a specific robot feature is in contact with a specific obstacle feature is a $(d-1)$ dimensional subset, or hypersurface, in the $d$ dimensional configuration space. Each combination of a robot feature and an obstacle feature defines such a hypersurface. If the total number of features (the complexity) of the robot and the obstacles are $O(1)$ and $m$ respectively, then the total number of hypersurface equals $O(m)$. 
The contact hypersurfaces can obviously intersect with each other. A point on the intersection of two hypersurfaces corresponds to a double contact of the robot. A point of intersection $j$ hypersurfaces corresponds to a $j$-fold contact of the robot with the obstacles. Such a $j$ fold contact defines an $(d-j)$ dimensional subset of the configuration space. Now an $d-1$ fold contact determines a hypersurface in the configuration space. The number of $d-1$ fold contact is $O(m^{d-1})$. Now we can state a following theorem.

\begin{theorem}
Let $A$ be a constant complexity robot with $d$ degrees of freedom and let $O$ be an obstacle with total face complexity $O(l^n)$ where $l$ is the number of hyperplanes surrounding the obstacle and $n$ is the dimension of the workspace. Then the $d-1$ dimensional face complexity of the configuration space obstacle is $O(l^{n^{d-1}})$.
\end{theorem}

From this we can draw a relationship between workspace obstacle surface area and configuration space obstacle hypersurface area. Since total surface area is dependent on number of faces, therefore it is more probable, that the obstacles having larger surface area in the workspace will also have a larger hypersurface area in the configuration space compared to obstacles that have smaller surface area in the workspace, within same tight bounding volume.

\section{The OBPRM Algorithm}\label{obprm_algo}
We first outline the OBPRM algorithm. Here, only the obstacles surface node generation phase is outlined. For 
complete algorithm descriptions and its various other details, see \cite{choset}, \cite{amato}.
\begin{algorithm}[!htb]
  \caption{OBPRM: Node Generation on C-Obstacles}
\begin{enumerate}
 \item  Determine a point $O$ (the origin) inside the C-obstacles, called the registration point.
 
 \item Select $N$ rays from origin $o$ with directions uniformly distributed in $C$-space.
 \item For each ray, use binary partition algorithm to determine a point on the boundary of C-Obstacles that lies on $\mathbb{X}_{free}$.
\end{enumerate}
\end{algorithm}
\begin{figure}[!h]
 \centering
 \def\svgwidth{150pt}
 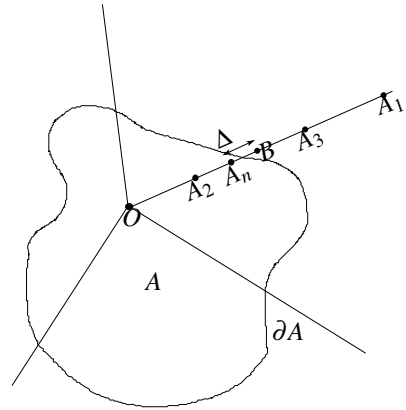
 \caption{OBPRM:Uniformly Distributed Surface Points Generation}
 \label{OBPRM_Fig}
\end{figure}  

In the first step, a point within the obstacle is selected as the origin, called the registration point. One would like to select this point near the centroid of the obstacle to generate almost uniformly distributed surface points. The rays in Step $2$ can be selected in a random manner, or, in a way such that a regular partition of the configuration space is obtained. Then, a binary partitioning algorithm in step $3$ can be carried out until a minimum step size is reached, or, a free point on the 
surface is reached. Generally, for all practical purposes, the number of iteration is limited to a fixed finite 
integer. In the final stage if the outcome of the binary space partitioning is a point within $\mathbb{X}_{free}$, then it is added to a graph structure, otherwise, that point is discarded. 

The node generation algorithm in the narrow passage region clearly depends on the shape of the configuration space obstacles. If obstacles are roughly spherical and the centroid is chosen as origin, then the resulting nodes may obtain a uniform distribution. Also a ray may intersect a boundary multiple times. In which case the node generation process may end up with a node very close to a local cavity instead of being in a true critical narrow passage region. For more details and implementation aspects see \cite{amato}.
\section{Definitions}\label{def}
Geometric probability studies invariant properties of sets of geometric objects undergoing various transformations. When considering
straight lines, pairs of points or triangles in space, one calculates the invariant measure on the variety of straight lines etc. This is 
initially started by Buffon with his famous needle problem \cite{buffon}, Santalo \cite{santalo}, Crofton \cite{crofton} and later developed by Rota \cite{rota}. 

Following are some definitions and theorems which will be usefull for calculating the success probability of OBPRM. For necessary details see \cite{rota}.

A partially ordered set, $L$ is called a lattice, if $\forall x,y \in L$, the least upper bound $x \vee y$ and the greatest lower bound
$x \wedge y$ also exists in $L$.

A lattice is distributive, if $\forall x,y,z \in L$,
\begin{itemize}
 \item[1.] $x\vee(y\wedge z)= (x\vee y)\wedge(x\vee z)$
 \item[2.] $x\wedge(y\vee z)= (x\wedge y)\vee(x\wedge z)$
\end{itemize}

A valuation on a lattice $L$ defined as a real valued function, which satisfies,$\forall A, B \in L$,
\begin{itemize}
 \item[1.] $\mu(A \cup B) = \mu(A) + \mu(B) - \mu(A \cap B)$
 \item[2.] $\mu(\emptyset) = 0$
\end{itemize}
Using Groemer's Integral theorem, a valuation on lattice can be extended to the valuation on distributive lattice.

Let $S$ is a non-empty set of $n$ elements. $P(S)$ is the set of all subsets of $S$, having set inclusion as a partial ordering. 
A chain in $P(S)$ is a linearly ordered subset where, for every pair $(x,y)$, either $x \leq y$ or $y \leq x$. An antichain is a subset $A$,
such that if $x,y \in A$ then $x \nleq y$, and $y \nleq x$. Denote $P_k(S)$ as an antichain consisting of all elements of $P(S)$ of rank $k$.
A simplicial complex defined as a subset
$A \in P(S)$, such that if $x \in A$, and $y \leq x$, then $y \in A$. A simplicial complex, if it contains a unique maximal element is called a simplex. For 
$x \in P(S)$, $\bar{x}$ is the simplex whose maximal element is $x$.

Next we state a theorem, as in \cite{rota}
\begin{theorem}
 Every valuation $\mu$ on the distributive lattice $L(S)$ of all simplicial complexes is uniquely determined by the values $\mu(\bar{x})$, $x \in P(S)$. The
 values of $\mu(\bar{x})$ can be arbitrary assigned.
\end{theorem}

We define $\mu_i(\bar{x}) = |\bar{x} \cap P_i(S)|$ and it can be extended to all of $L(S)$. Note that these valuations can also be expressed in terms of 
symmetric functions, as can be found in \cite{rota}.

Denote, $\mathbb{K}^n$ as the set of all compact convex subsets of $\mathbb{R}^n$. A finite union of compact convex sets is called a polyconvex set.
Family of polyconvex sets in $\mathbb{R}^n$ is a distributive lattice, denoted as $polycon(n)$. Note that all configuration space obstacles
can be classified as polyconvex sets.

Next, we state an important theorem, due to Groemer \cite{rota}
\begin{theorem}
 A convex continuous valuation $\mu$ on $\mathbb{K}^n$ admits an unique extension to a valuation on the lattice $polycon(n)$
\end{theorem}

A valuation $\mu$ defined on polyconvex sets in $\mathbb{R}^n$ is said to be invariant if $\mu(A) = \mu(gA)$, $\forall g \in \mathbb{E}_n$, and $\forall
A \in polycon(n)$, Here $g$ is a motion or transformation under the euclidean group  $\mathbb{E}_n$ is on $\mathbb{R}^n$.

Next, define the partially ordered set of all linear varieties in $\mathbb{R}^n$ as affine varieties in $\mathbb{R}^n$ or $\rm{Aff(n)}$. The subset of $\rm{Aff(n)}$
consisting of all elements of rank $k$, that is, all linear varieties of dimension $k$ is called the affine Grassmannian or $\rm{Graff(n,k)}$. For example,
$\rm{Graff(2,1)}$ denotes the set of all straight lines in $\mathbb{R}^n$. 

Define a measure $\lambda_k^n$ on $\rm{Graff(n,k)}$ that is invariant under the group $\mathbb{E}_n$. Then according to the Hadgwiger formula, 
$\mu_{n-k}(A) = C_k^n\int_{V \in \rm{Graff(n,k)}}\mu_0(A \cap V)d\lambda_k^n(V)$, $\forall A \in polycon(n)$. Here $\mu_0$ is called the Euler characteristic
and $C_k^n$ is some constant.

Next we state Sylvester's theorem, \cite{sylvester}.
\begin{theorem}
Let $K \subseteq L$ be compact convex sets. Suppose $L$ is of dimension $n$. The conditional probability that a linear
variety of dimension $k$ shall intersect $K$, given that it intersects $L$, is given by, $\mu_{n-k}(K)/\mu_{n-k}(L)$. 
\end{theorem}
Next we state the following result, due to Hadgwiger ch.7 \cite{rota}, for the lattice of polyconvex sets $polycon(n)$.
\begin{theorem}
 The valuations $\mu_0,\mu_1,\ldots,\mu_n$ form a basis for the vector space of all convex continuous rigid motion invariant valuations defined 
 on polyconvex sets in $\mathbb{R}^n$.
\end{theorem}

Define the volume of the unit ball in $\mathbb{R}^n$ as $\omega_n = \frac{\pi^{n/2}}{\Gamma((n/2)+1)}$. The valuations $\mu_i$ of unit ball $B_n$ is given as,
for $0\leq i \leq n$, $\mu_i(B_n) = {n \choose i}\frac{\omega_n}{\omega_{n-i}}$.

Finally, we state the principal kinematic formula.
\begin{theorem}
 For all $A,K \in polycon(n)$,\\
 \begin{equation}
 \int_{\mathbb{E}_0} \mu_0(A \cap gK)dg = \sum_{i=0}^{n}{n \choose i}^{-1}\frac{\omega_i \omega_{n-i}}{\omega_n}\mu_i(A)\mu_{n-i}(K) 
 \end{equation}
 $\forall g \in \mathbb{E}_n$.
\end{theorem}
\section{Probabilistic Analysis of OBPRM}\label{analysis}

Here, we present the probabilistic analysis of OBPRM algorithm. Although the use of the term probabilistic analysis is certainy not appropriate here, because the aim is not to determine the average case analysis of OBPRM algorithm. Instead we are interested in calculating the probability of the event that after $N^{th}$ iteration, for a single ray, a point lying in $\mathbb{X}_{free}$ can be obtained.

Let the $i^{th}$ obstacle be denoted by $A_i$ and its boundary by $\partial A_i$. In general motion planning scenarios, often the configuration space obstacles are curved and can be represented as semi algebraic sets. This makes any analysis extremely difficult and one has to compromise in order to obtain any meaningful result. Therefore we assume that the obstacles are arbitrarily close approximated by intersections of a finite
number of hyperplanes in $\mathbb{R}^n$. To obtain the analysis we choose a configuration space obstacle randomly and here onwards we will denote it as $A$.

Let $O$ a point within the obstacle, which acts as origin and $\overline{OA_1}$, is one such example of a ray coming out of origin $O$ (see Fig \ref{OBPRM_Fig}).

Let the length of the line segment $\overline{OA_1}$ be $l$. Let, after $N$ steps, the length of the smallest binary partition, be denoted as $d_N$. Now,
after each binary segment division of the line segment, the size of the resultant smallest partition, forms a sequence, $(l/2,l/4,\ldots,l/2^N)$. Now, similar to a Monte-Carlo approach, the partitioning process can stop after a certain smallest partition size is reached, or else, similar to the Las-Vegas approach it can continue until a point in $\mathbb{X}_{free}$ is obtained. Both the processes are equivalent in terms of calculating the success probability. We choose a general Monte-Carlo setting and assume that the algorithm
stops when a minimum step size, denoted as $\Delta$ is reached. Clearly, the algorithm stops after $N$ binary division if, $d_N \leq \Delta$, or when $N \geq \log_2 (l/\Delta)$. This implies that, after at least $\log_2(l/\Delta)$ iterations, we may expect a point $B$ (see Fig \ref{OBPRM_Fig}),
which is in $\mathbb{X}_{free}$, such that the smallest partition size is $\Delta$.

The question we are addressing is, what is the probability that the final point $B$ will be in $\mathbb{X}_{free}$. Equivalently, we may look for the probability of the event that after $N$ iterations $\overline{A_NB}$ must intersect the boundary $\partial A$. Because this guarantees the existence of the free point $B \in \mathbb{X}_{free}$. 

Now the obstacles may have arbitrary shapes, the origin can be any point inside the obstacle, and the location where $\overline{A_NB}$ may intersect $\partial A$ is arbitrary. If we sample random line segment  $\overline{A_NB}$, for each intersection of $\partial A$ with random segment, one can think of a hypothetical origin $O$ from which a ray has been cast and lead to the random intersection with boundary $\partial(A)$. We therefore reformulate the problem as, if we drop randomly a line segment $\overline{A_NB}$, what is the probability that it will hit $\partial A$ ? Notice the similarity of the problem with the famous buffon's needle problem \cite{buffon}.

\begin{figure}[htb]
 \centering
 \def\svgwidth{150pt}
 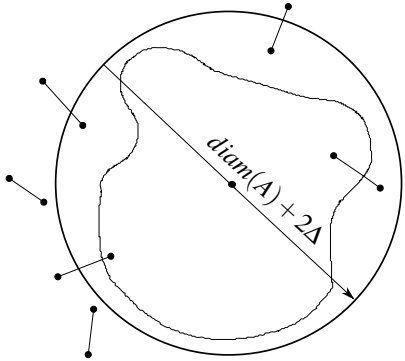
 \caption{OBPRM:Line Segments under different motions $g \in \mathbb{E}_n$}
 \label{OBPRM_Fig_2}
\end{figure}

This problem can be viewed as, finding the measure of the set of motions $g\in \mathbb{E}_n$, such that, $\partial A \cap g(A_NB)$ is non-zero.
This is a standard
problem in geometric probability theory, and using principal kinematic formula, we can derive the corresponding success probability, see Fig \ref{OBPRM_Fig_2}.
We assume that, the configuration space obstacles are created by the intersections of finite number of halfspaces. They all belong to $polycon(n)$. 
For simplicity,
we assume that the obstacles are convex shaped. If not one can use the convex hull formed by the vertices of the configuration space obstacles. This is not a serious restriction as the formulas are equally applicable for non convex sets also.
Now, since the family of polyconvex sets in $\mathbb{R}^n$ is a distributive lattice, therefore we can define valuations on them, as defined in Section \ref{def}.

Denote $K$ as the line segment $\overline{A_NB}$. Now, a direct use of principal kinematic formula gives us,
\begin{equation}\label{1}
   \int_{E_n} \mu_0 (\partial A \cap g k) dg = \sum_{i=0}^n {n \choose i}^{-1} \frac{\omega_i \omega_{n-i}}{\omega_n} \mu_i (\partial A) \mu_{n-i}(k)
\end{equation}

where the left hand side indicates the measure of the set of different motions in $\mathbb{E}_n$ such that, when applied to convex line segment $K$,
that is, $g(K)$, it intersects $\partial A$. 

Now consider also a closed ball $B_n$, whose diameter is $d = diam(A) + 2\Delta$. For the same line segment $K$, the measure of the set of motions in 
$\mathbb{E}_n$, for which $K$ intersects $B_n$ is defined as,

\begin{equation}\label{1}
   \int_{E_n} \mu_0 (B_d \cap g k) dg = \sum_{i=0}^n {n \choose i}^{-1} \frac{\omega_i \omega_{n-i}}{\omega_n} \mu_i (B_d) \mu_{n-i}(k)
\end{equation}
Since $A \subseteq B_n$, following a similar argument behind Sylvester's theorem, defined in section \ref{def}, we define the conditional probability that a line segment
$K$ will intersect $\partial A$, given that it intersects $B_n$, as
\begin{eqnarray*}
  P_r &=& \frac{\int_{E_n} \mu_0(\partial A \cap g k) dg}{\int_{E_n} \mu_0 (B_d \cap g k) dg} \\
   &=& \frac{\sum_{i=0}^n {n \choose i}^{-1} (\frac{\omega_i \omega_{n-i}}{\omega_n}) \mu_i (\partial A) \mu_{n-i}(k)}{\sum_{i=0}^n {n \choose i}^{-1}(\frac{\omega_i \omega_{n-i}}{\omega_n}) \mu_i (B_d) \mu_{n-i}(k)} 
\end{eqnarray*}
Since, $\mu_0(K) = 1$ (The Euler characteristic), $\mu_1(K) = \Delta$ and for all $2 \leq  i \leq n$, $\mu_i(K) = 0$, therefore,

\begin{equation*}
  P_r = \frac{{n \choose i}^{-1}  (\frac{\omega_{n-1} \omega_1}{\omega_n}) \mu_{n-1} (\partial A) \mu_1 (k)+ \omega_0 \mu_n (\partial A) \mu_0(k)} 
{{n \choose i}^{-1}  (\frac{\omega_{n-1} \omega_1}{\omega_n}) \mu_{n-1}(B_d)\mu_1 (k)+ \omega_0 \mu_n (B_d) \mu_0(k)} 
\end{equation*}
Since, $\omega_0 =1$ and $\mu_n(\partial A) = 0$
\begin{equation*}
  P_r = \frac{(\frac{\omega_{n-1} \omega_1}{\omega_n}) \mu_{n-1} (\partial A) \mu_1 (k)+0}{(\frac{\omega_{n-1} \omega_1}{\omega_n}) \mu_{n-1} (B_d) \mu_1 (k)+ \mu_n (B_d)}
\end{equation*}
or,
\begin{equation*}
  P_r = \frac{(\frac{\omega_{n-1} \omega_1}{\omega_n}) \mu_{n-1} (\partial A)\Delta}{(\frac{\omega_{n-1} \omega_1}{\omega_n}) \mu_{n-1} (B_d) \Delta+ \mu_n (B_d)}
\end{equation*}
Now, 
\begin{eqnarray*}
\mu_n (B_d)={n \choose i}\frac{\omega_n}{\omega_0}=\omega_n 
\end{eqnarray*}
$\mu_{n-1} (B_d) ={n \choose i}\frac{\omega_n}{\omega_1}=\frac{n \omega_n}{d}$

Now, for a given $\Delta$, let, $\frac{\omega_{n-1}\omega_1}{n \omega_n}=\alpha$
\begin{eqnarray*}
  P_r &=& \frac{\alpha \mu_{n-1}(\partial A) \Delta}{\alpha \frac{n\omega_n}{2d}\Delta +\omega_n } \\
   &=& \frac{\alpha}{\omega_n}\left[
                           \begin{array}{c}
                             \frac{ \mu_{n-1}(\partial A) \Delta}{\frac{\alpha. n}{2d} \Delta +1} \\
                           \end{array}
                         \right]
\end{eqnarray*}

Therefore, the success probability 
\begin{equation*}\label{2}
    P_r= \frac{\alpha}{\omega_n}\left[
                           \begin{array}{c}
                             \frac{ \mu_{n-1}(\partial A) \Delta}{\frac{\alpha. n}{2d} \Delta +1} \\
                           \end{array}
                         \right]
\end{equation*}
Since we assume that the obstacles are either convex shaped, or can be approximated as convex hull of obstacle vertices, this success
probability can be thought of as a conservative estimate of the actual one. 
This indicates that for constant $\Delta$, and $d$, if the $\mu_{n-1}(\partial A)$ increases, the success probability also increases. $\mu_{n-1}(\partial A)$
indicates surface area of $n$ dimensional obstacle $A$. Also,
if the obstacles enlarges, then $d$ also increases. In that case, we cannot say in general about the nature of the probability of success. Note that, 
for a fixed $d$, since we assume that the obstacle is actually made up of finitely many intersections of halfspaces, one can relate
the concept of box-dimension \cite{} with the success probability. 
\section{Simulation Results}\label{Results}

\begin{figure}[!h]
  \subfloat[]{\includegraphics[height=4cm,width=4cm]{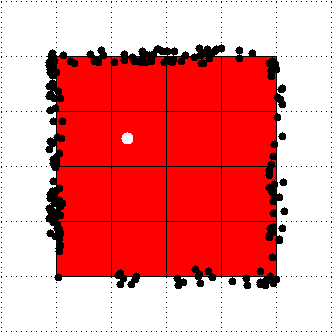}}\hspace*{0.5cm}
  \subfloat[]{\includegraphics[height=4cm,width=4cm]{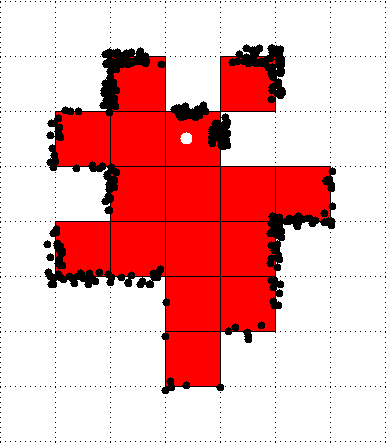}}\\
  \subfloat[]{\includegraphics[height=4cm,width=4cm]{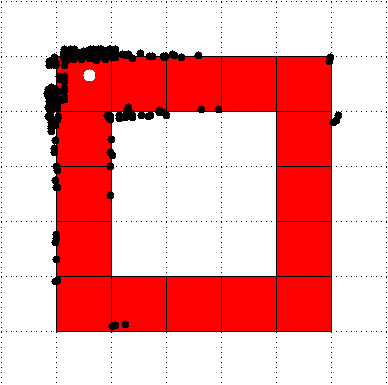}}\hspace*{0.5cm}
  \caption{Figures showing free configuration after $200$ iterations.}
  \label{fig:Result_1}
\end{figure}

We investigate the validity of the result in a simplistic $2$-D situation as well as in a complex $3$D environment. In the $2$D case, the configuration space is $\mathbb{R}^2$. We consider $3$ different type of obstacles and try to generate free points on their surfaces as uniformly as possible. In each case, we choose $\Delta = 1$ unit, number of rays $N = 200$. Each obstacle has the same area but different perimeter. The objective is to see if the success probability increases with the increase of perimeter. Note that in all three cases the diameter of obstacles are not exactly equal. We believe that this will not have drastic effect on the simulation. We use a Pentium core i$7$ processor with $2$GB RAM. All the simulations is performed in a MATLAB/Simulink environment.
For each of the cases we did a $100$ Monte-Carlo simulations to obtain an average value of success probability. In Fig \ref{fig:Result_1}, case $(1)$ the average value of the 
success probability is $0.45$, while in case $(2)$, it is $0.55$ and in case $(3)$, it is $0.63$. This result is consistent with the fact that
Perimeter (Case 1) $<$ Perimeter (Case 2) $<$ Perimeter (Case 3).

For a complex scenario we choose a room with different test objects such as a piano. The intention was to create the situation of famous piano movers problem, as can be found in \cite{reif}. All the models are polygonal models. We choose two different models of the piano, one having a lager number of vertices and polygonal faces than the other. Following figures \ref{fig:Environment} and \ref{fig:Piano} shows the environment and both the high and low polygonal piano. Both the model is within unit bounding volume. Although not calculated directly, according to Section \ref{Cspace Obstacles}, we can assume that low polygonal piano model will create configuration space obstacle having lesser surface area. We use a modified depth buffer and occlusion query based collision detection scheme, which is fast and efficient. For the low polygonal piano model we found probability of generating free configurations on the obstacle surface is $0.14$, while for the high polygonal piano model, it is $0.24$. This value is the average of $100$ iterations.
\begin{figure}[!h]
  \centering
  \includegraphics[height=6cm,width=8cm]{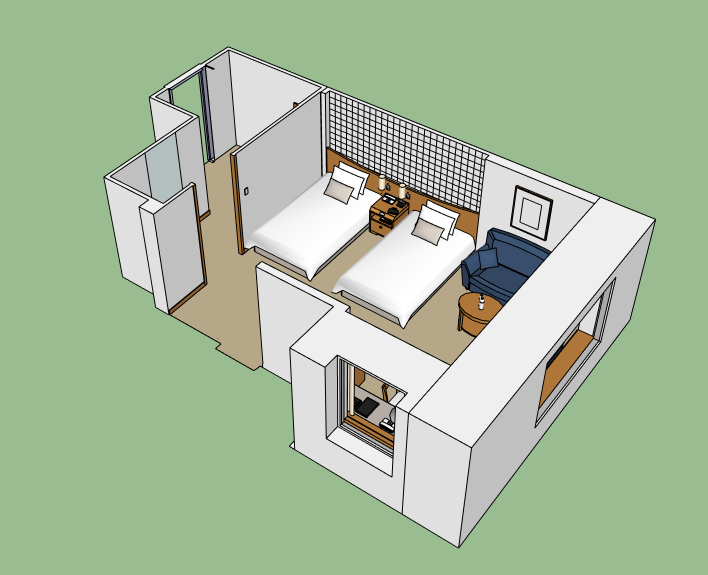}
  \caption{Figure showing the environment model}
  \label{fig:Environment}
\end{figure}

\begin{figure*}[!htb]
  \centering
  \subfloat[]{\includegraphics[height=6cm,width=7cm]{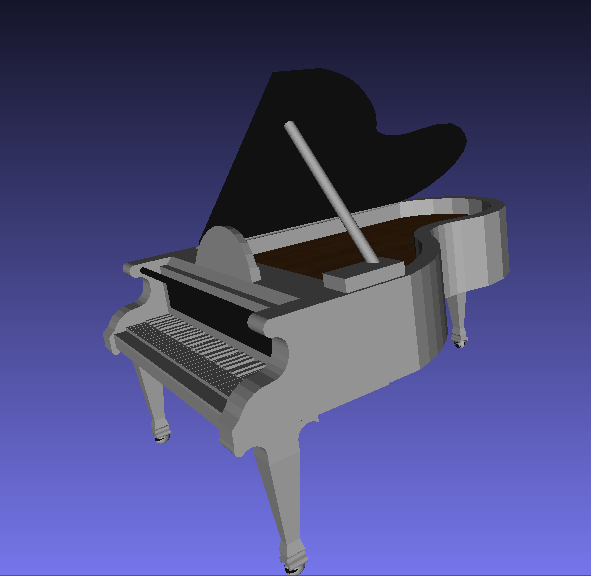}}\hspace*{1cm}
  \subfloat[]{\includegraphics[height=6cm,width=7cm]{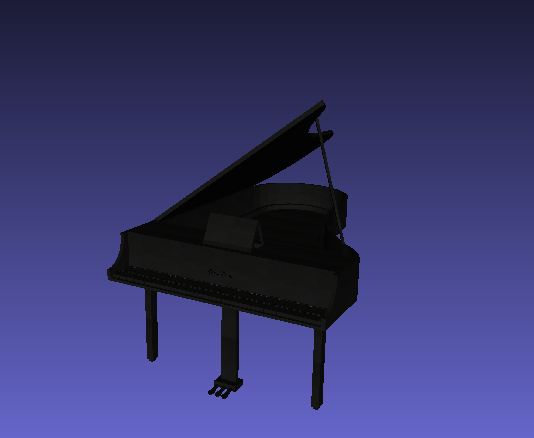}}
  
  \caption{Polygonal modeles of Pianos. a)High polygonal model with $12862$ faces and $19067$ vertices b) Low polygonal model with $7864$ faces with $9456$ vertices}
  \label{fig:Piano}
\end{figure*}
\begin{figure*}[!htb]
  \centering
  \subfloat[]{\includegraphics[height=6cm,width=7cm]{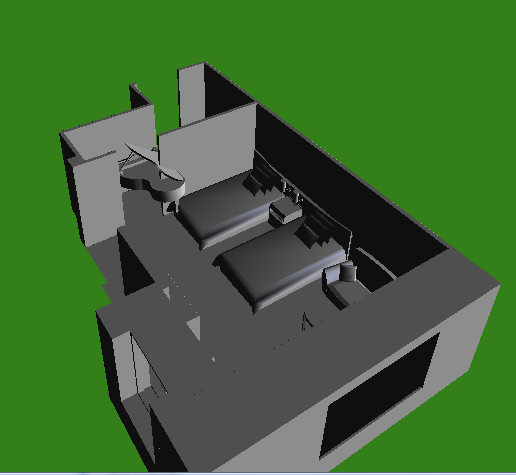}}\hspace*{1cm}
  \subfloat[]{\includegraphics[height=6cm,width=7cm]{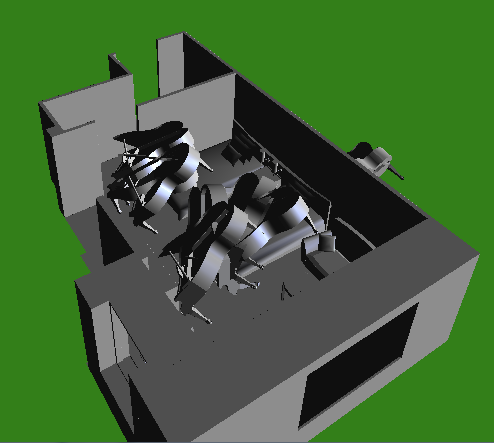}}\\
  \subfloat[]{\includegraphics[height=6cm,width=7cm]{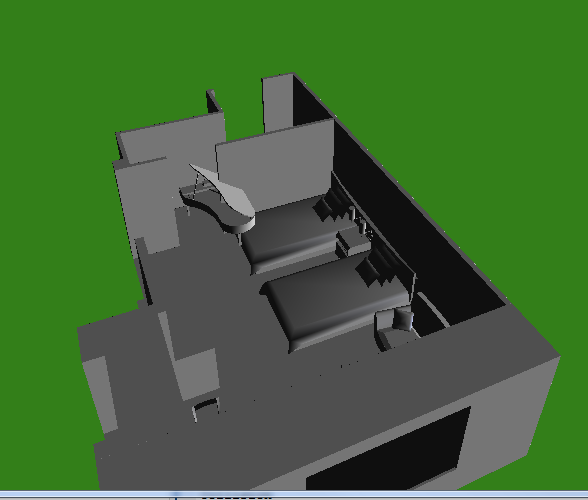}}\hspace*{1cm}
  \subfloat[]{\includegraphics[height=6cm,width=7cm]{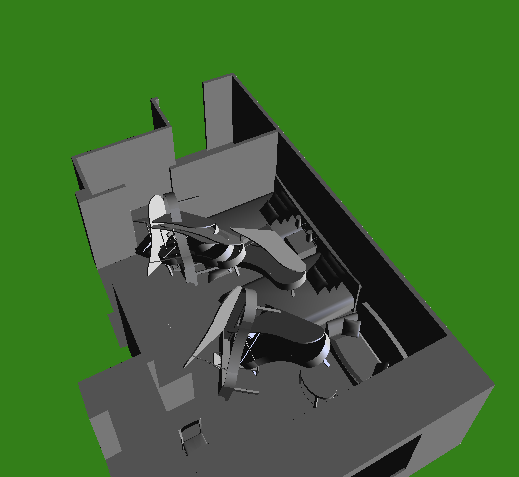}}
  \caption{a) High polygonal piano in a colliding position, which is the registration point b) Generated free configurations after 50 iterations c) Low polygonal model piano in a colliding position d) Genrated Configurations after 50 iterations}
  \label{Figure_Incremental}
\end{figure*}

\section{conclusion}\label{Conclusion}
In this paper, we have shown that the success probability of OBPRM algorithm is proportional to the 
surface area of $n$ dimensional configuration space obstacles. It is also practically shown using
simplistic scenarios in $\mathbb{R}^2$ through simulation. It would also be interesting to investigate such results in various other high dimensional situation. 

Such applications of geometric probability results are new in the literature. Using the rich theory of stochastic geometry, we believe that many other sampling based algorithms such as, randomized bridge builder (RBB) or toggle PRM also can be analyzed in future work in this direction.


\begin{thebibliography}{99}
\bibitem{perez}
T. Lozano-Perez., Spatial planning: a configuration space approach,"\emph{IEEE Transaction on Computers}, vol. 32, no. 2, pp. 108-120, 1983.
\bibitem{reif}
Reif, J. H., Complexity of the mover's problem and generalization," \emph{In: Proc. of 20th Annual Symposium on Foundations of Computer Science}, San Juan, Puerto Rico, pp. 421-427, 1979.
\bibitem{kavraki}
Kavraki, L. E., Svestka, P., Latombe, J. C. and Overmars, M. H.,
Probabilistic roadmap for path planning in high dimensional
configuration spaces," \emph{IEEE Transaction on Robotics and Automation},
vol. 12, no. 4, pp. 566-580, 1996.

\bibitem{lavalle}
Lavalle, S. M., Rapidly exploring random tree," \emph{Technical
Report TR 98 -11, Computer Science Department}, Iowa State
University, 1998.

\bibitem{choset}
Choset, H. et al., Principles of robot motion: theory algorithms and
implementations," \emph{MIT Press}, Cambridge, Massachusetts, 2005.

\bibitem{carpin}
Carpin, S., Randomized motion planning - a tutorial," \emph{International
Journal of Robotics and Automation}, vol. 21, no. 3, pp. 184-196, 2006.

\bibitem{amato}
Amato, N. M., Bayazit, O. B., Dale, L. K., Jones, C. and Vallejo, D., 
OBPRM: An obstacle-based PRM for 3D workspaces," \emph{In: Proc. 3rd
Workshop Algorithmic Foundations on Robotics}, pp. 155-168, 1998.
\bibitem{kavaraki_2}
Kavraki, L. E., Kolountzakis, M. N. and Latombe, J. C.,Analysis of probabilistic roadmaps for path planning," \emph{IEEE Transactions on Robotics and Automation}, vol. 14, no. 1, pp. 166-171, 1998
\bibitem{boor}
Boor, V., Overmars, M. H. and Van der Stappen, A. F., The Gaussian
sampling strategy for probabilistic roadmap planners," \emph{
Proceedings of the 1999 IEEE International Conference on Robotics
and Automation}, pp. 1018-1023, 1999.
\bibitem{Zheng}
Sun, Z., Hsu, D., Jiang, T., Kurniawati, H. and Reif, J. H., Narraow
Passage Sampling for Probabilistic Roadmap Planning," \emph{IEEE
Transaction on Robotics} vol. 21, no. 5, 1105-1115, 2005.
\bibitem{mulmulle}
Ketan Mulmuley, Computational Geometry - An Introduction through Randomized Algorithms," \emph{Prentice Hall}, 1994.
\bibitem{buffon}
Feller, W., Introduction to Probability Theory and Its Applications, " \emph{New York: Wiley Publishers, 68}, vol. 2, 2014.
\bibitem{santalo}
Luis A. Santalo., Integral Geometry and Geometric Probability," \emph{Addison-Wesley}, Reading, MA, 1976
\bibitem{crofton}
Crofton, M. W., Geometrical Theorems relating to mean values, "\emph{Proc. London Math. Soc}, no. 8, pp. 304-309, 1868.
\bibitem{rota}
Klain, D. A. and Rota, G. C., Introduction to geometric probability," \emph{Cambridge University Press}, 1997.
\bibitem{sylvester}
Sylvester, J. J., On a funicular solution of Buffon's 'problem of the needle' in its most general form, " \emph{Acta Math.,}, no. 14, pp 185-205, 1890.

\end{thebibliography}
\end{document}